\documentclass[runningheads]{llncs}
\usepackage{graphicx}
\usepackage{booktabs}
\usepackage{amsmath,amssymb}
\usepackage{float}
\graphicspath{{./}}
\begin{document}
\title{Reverse Spatio-Temporal Disease Progression Modelling}
\titlerunning{Reverse Disease Progression Prediction}
\author{Ulugbek Shernazarov\inst{1}$^{*}$ \and
Moucheng Xu\inst{2}$^{*,\dagger}$ \and
Inomjon Ramatov\inst{3}}

\authorrunning{U. Shernazarov et al.}

\institute{
Télécom SudParis, Institut Polytechnique de Paris, France\\
\email{ulugbek.shernazarov@telecom-sudparis.eu}
\and
University College London, UK
\and
Digital Technologies and Artificial Intelligence Development Research Institute (AIRI), Uzbekistan
}
\maketitle

\begingroup
\renewcommand{\thefootnote}{\fnsymbol{footnote}}
\footnotetext[1]{These authors contributed equally to this work.}
\footnotetext[4]{This work started when Moucheng was with UCL; he has now moved to Medtronic}
\endgroup

\begin{abstract}
Deep learning-based spatio-temporal disease progression models commonly overlook the incubation period of progressive diseases, limiting the use of those models in early interventions, which are vital for not easily reversible diseases such as Alzheimer's. This is because, the existing deep learning based longitudinal disease-progression models are almost always run \emph{forward}: from an observed baseline they predict future decline. In many clinical settings, however, imaging begins only after pathology is suspected or already visible, the earlier, healthier patient-specific reference was never acquired. To address this, we propose to study \emph{reverse disease progression prediction}: given later diseased anatomy, reconstruct the unobserved healthier anatomy that preceded it. We use a two-stage model in which a frozen 3D vector-quantised autoencoder defines a compact discrete latent space, while a Neural Ordinary Differential Equation (ODE) learns continuous-time dynamics in that space. A recurrent encoder reads late observations in reverse temporal order, initialises the latent state, and the ODE is integrated backwards across the trajectory. On a controlled Morpho-MNIST benchmark with a sinusoidal perturbation, our model successfully recovered the unseen previous states from later observations of the non-monotonic trajectory. On longitudinal brain MRIs from Alzheimer’s Disease Neuroimaging Initiative, at the task to recover the previous unseen trajectory towards healthy states of the patients from observed later diseased states, our model outperforms the baselines that uses copy-nearest and mean-observed, with positive disease-reversal scores  in every diagnostic stratum. We hope that our work can provide insights and tools towards discovering the incubation periods from single-shot scans, and developing early interventions of diseases based on imaging.

\end{abstract}

\section{Introduction}
Existing spatio-temporal disease progression modelling approaches~\cite{RAVI2022102257,Pug_Enhancing_MICCAI2024} focus on the forward direction: given a baseline scan, a model predicts future anatomy. Recent generative methods make such temporal synthesis feasible in 3D by compressing volumes into discrete or latent representations and then learning temporal evolution~\cite{oord2017neural,esser2020taming,rombach2021highresolution,yoon2022sadm,zhao2025vqgan}. This direction limits the uses of those models in early interventions, especially on discovering the incubation periods, and detecting early stages from the imaging. 


Existing counterfactual and normative approaches estimate healthy appearance by comparing a patient with population distributions, anomaly maps, brain-age deviations, or disease-progression stages~\cite{baur2018deep,bintsi2020patchbased,marquand2019conceptualizing,young2024datadriven}. These methods are valuable, but they do not reconstruct the same subject's own earlier anatomy from that subject's later scans. We formulate this missing-baseline setting as \emph{reverse disease progression prediction}: given later diseased states, synthesize the unobserved earlier state. The task is clinically meaningful for progressive, largely irreversible processes such as neurodegeneration, where earlier time is a reasonable proxy for healthier anatomy, but it is also technically fragile because slowly changing registered anatomy makes copying the closest observed scan a deceptively strong baseline.

In order to address the problem of \emph{reverse disease progression prediction}, we propose a two-stage latent Neural ODE that learns continuous-time reverse dynamics in a frozen discrete representation. The same formulation supports irregular visit schedules and, when only one late scan is observed, single-scan counterfactual inference. Our approach is evaluated on two datasets, a simulated dataset with non-monotonic dynamics of morphological deformations, and another real dataset from Alzheimer’s Disease Neuroimaging Initiative (ADNI). Our experimental results show that our model can recover the unseen trajectory preceding  to the observed states, implying potential clinical utility to discover incubation periods to help with early detections and early interventions. 


\section{Related Work}
Vector-quantized representation learning and VQ-GANs compress images into tractable discrete code grids~\cite{oord2017neural,esser2020taming}, while latent diffusion performs synthesis in a compressed latent space~\cite{rombach2021highresolution}. For longitudinal medical imaging, sequence-aware diffusion models and 4D-VQ-GAN synthesize follow-up scans or scans at arbitrary future time points~\cite{yoon2022sadm,zhao2025vqgan}. Our work reverses the inference direction: instead of forecasting from an observed baseline, it reconstructs a subject-specific earlier baseline from later observations. Neural ODEs parameterize continuous-time latent dynamics~\cite{chen2018neural}; latent ODEs are especially appealing for irregular clinical visits~\cite{rubanova2019latent}. However, slowly changing anatomy can make a near-zero vector field appear competitive, so we explicitly test for non-trivial dynamics.

\section{Method}
\subsection{Problem Formulation and Architecture}
Let $\{\mathbf{x}_{t_i}\}_{i=0}^{T-1}$ be a subject sequence ordered from earliest to latest visit. During reverse prediction, only the final $n_{\mathrm{obs}}$ scans are observed, and the target is $\mathbf{x}_{t_0},\ldots,\mathbf{x}_{t_{T-n_{\mathrm{obs}}-1}}$. Stage 1 is a 3D VQ-GAN: an encoder $E$ maps a volume to a feature grid, a codebook $\mathcal{C}=\{\mathbf{e}_k\}_{k=1}^{K}$ quantizes it to a latent grid $\mathbf{z}$, and a decoder $D$ reconstructs the volume. Stage 1 is trained with reconstruction, adversarial, and perceptual terms~\cite{zhang2018unreasonable} following the VQ paradigm~\cite{oord2017neural,esser2020taming}, then frozen. We use $K=32$ for Morpho-MNIST and $64^3$ ADNI experiments, and $K=128$ for the $1\,$mm brain model.

\begin{figure}[H]
\centering
\includegraphics[width=\textwidth]{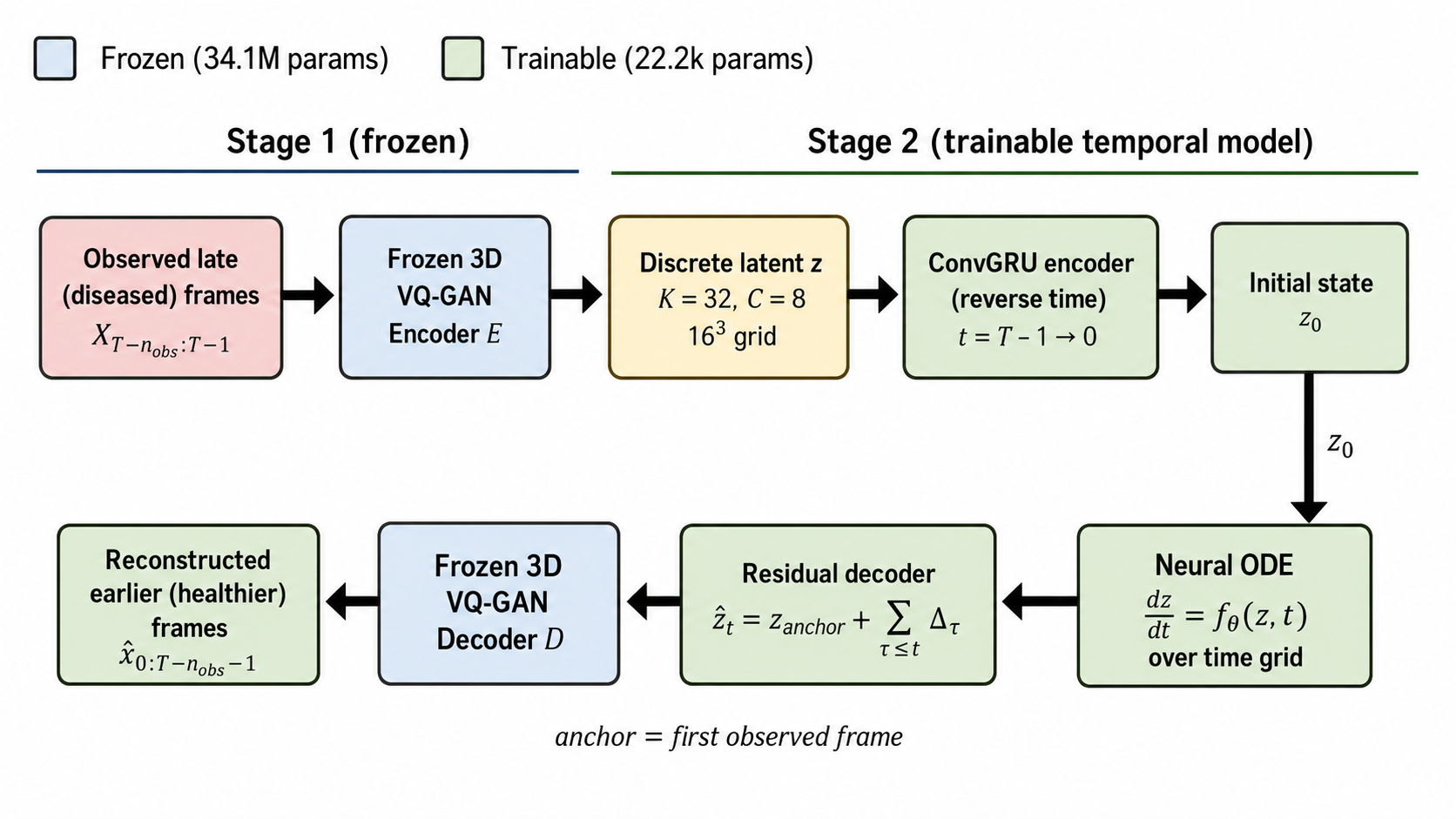}
\caption{Two-stage reverse-prediction framework. A frozen 3D VQ-GAN maps late scans to a discrete latent grid. A reverse-order recurrent encoder initializes the latent state, a Neural ODE integrates backward over the visit grid, and a residual latent decoder plus frozen image decoder reconstruct earlier frames.}
\label{fig:framework}
\end{figure}

Stage 2 learns the temporal model (Fig.~\ref{fig:framework}). Each observed scan is encoded to $\mathbf{z}_{t_i}$, a ConvGRU consumes the observed latent sequence in reverse order, and a Neural ODE integrates $d\mathbf{z}/dt=f_{\theta}(\mathbf{z},t)$ over the requested time grid. Only the ODE function and residual temporal head are trainable ($22{,}163$ parameters); the frozen Stage-1 model contains $34.1$M parameters. Stage 2 is trained for $20{,}000$ steps at learning rate $2\times10^{-4}$ in fp32, mixed precision produced NaNs therefore not used.

\subsection{Residual Reverse Dynamics and Single-Scan Inference}
\begin{figure}[H]
\centering
\includegraphics[width=\textwidth]{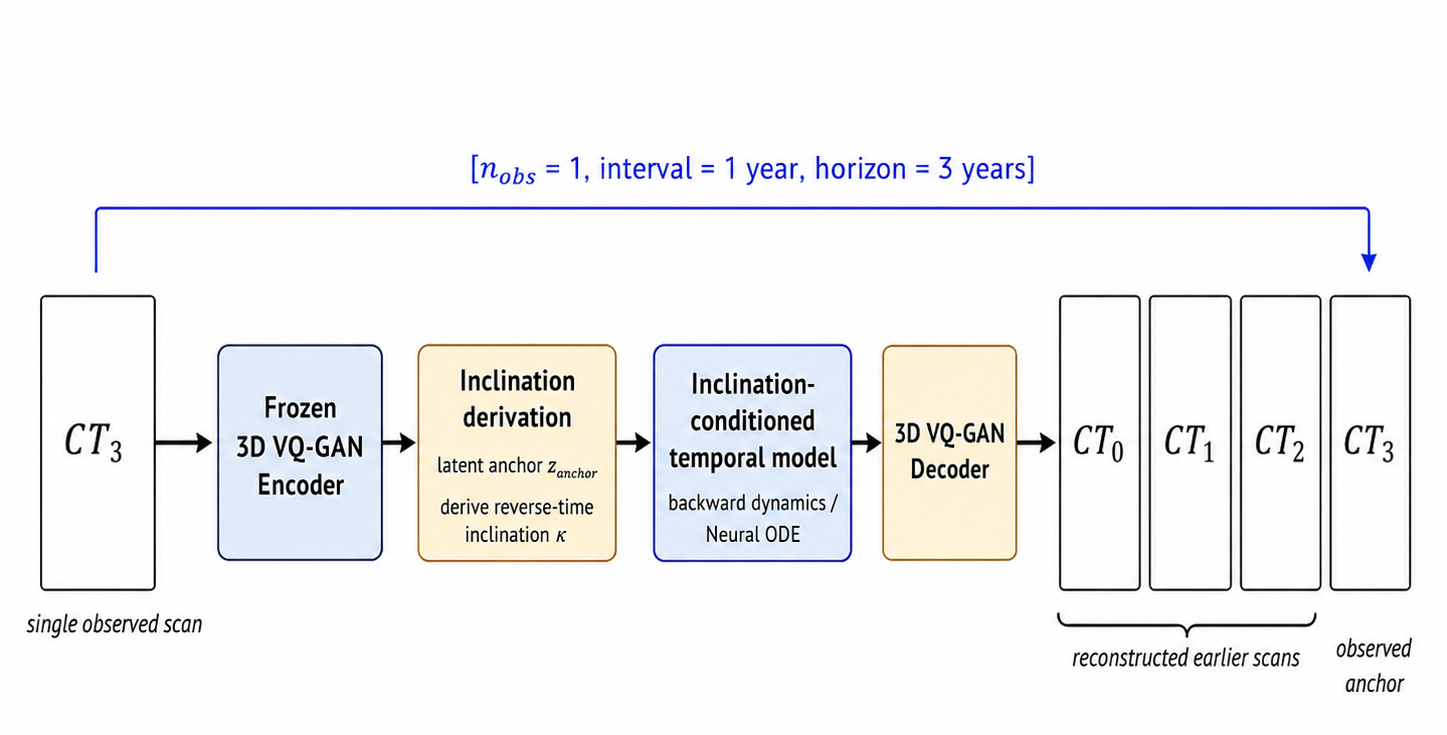}
\caption{Single-scan inference. From one observed late scan, the frozen encoder produces a latent anchor; an inclination-conditioned temporal model integrates backward over the requested interval and horizon; the frozen decoder reconstructs earlier scans.}
\label{fig:single}
\end{figure}

Rather than decoding each time point independently, the temporal head predicts latent increments $\Delta_t$ accumulated from the first observed anchor,
$\hat{\mathbf{z}}_t=\mathbf{z}_{\mathrm{anchor}}+\sum_{\tau\leq t}\Delta_{\tau}$.
This encourages patient-specific change relative to an observed state rather than memorization of absolute anatomy. The objective combines direct latent error and an intermediate frame-difference loss, $\mathcal{L}=0.2\mathcal{L}_{\mathrm{direct}}+\mathcal{L}_{\mathrm{inter}}$, optimized with Adam~\cite{kingma2014adam}. Correct masking is essential: the residual decoder is anchored at the first observed frame, not the hidden $t=0$ frame, and frame-difference losses are computed only between observed frames. Integer sequence indices are separated from the continuous times passed to the ODE solver, allowing integration on irregular clinical schedules. With $n_{\mathrm{obs}}=1$, the model initializes from one late scan; the learned vector field, conditioned on the latent anchor and requested horizon, supplies the reverse-time inclination (Fig.~\ref{fig:single}).

\section{Experiments}

\subsection{Dynamic Morpho-MNIST Dataset} 
We convert MNIST digits into $32^3$ pseudo-3D volumes and create 10-time-point sequences using a periodic perturbation whose amplitude follows a $1.5$-cycle sine wave. The deformation alternates thinning and thickening~\cite{castro2018morphomnist}, so static prediction cannot fit the full trajectory. 

\subsection{ADNI}
We use 382 T1-weighted MRI from ADNI 1 \cite{jack2008alzheimer}. Subjects with at least five valid dated visits and non-missing diagnosis are retained, and the latest five visits are used ($T=5$). Median visit days are $0,278,465,752,1120$, giving normalized times $[0,0.248,0.415,0.672,1.0]$. We use a subject-level $80/10/10$ split with 25 test subjects. 

For the preprocessing of ADNI 1 MRI, we use a register-then-strip pipeline: follow-up visits are rigidly aligned to the subject baseline with $6$ degrees of freedom and no scaling, preserving atrophy; the baseline is affinely registered to MNI152 with $12$ degrees of freedom; transforms are composed into one resampling; HD-BET skull-stripping is applied in MNI space~\cite{isensee2019hdbet,smith2002bet}; and a fixed brain box is cropped to $160\times188\times152$ voxels at $1\,$mm isotropic resolution. Registration uses mutual-information affine alignment equivalent to FLIRT~\cite{jenkinson2002flirt}. Without this step, longitudinal ``change'' is entangled with pose, field-of-view jitter, and resampling artifacts. We evaluate both unregistered $64^3$ resampling and the registered $1\,$mm MNI pipeline above. 

\subsection{Baselines}
Four simple baselines are implemented, including: 1) VQ copy-nearest, copy nearest codebooks then decoding; 2) VQ mean-observed, averages the codebooks of the observed frames, then decoding; 3) Pixel copy-nearest, copy nearest frames; 4) Pixel mean-observed, averages the observed frames. We report metrics only on the predicted past frames. 


\subsection{Metrics}
We report MSE, PSNR, and mid-slice SSIM~\cite{wang2004image} on predicted frames only. In addition, we define two task-specific diagnostics. The first is a disease-reversal score,
\[
\mathrm{Rev}
=
1-
\frac{\mathrm{MSE}(\hat{x}_t,x_t)}
{\mathrm{MSE}(x^{\mathrm{copy}}_t,x_t)},
\]
where $x^{\mathrm{copy}}_t$ is the nearest observed diseased scan copied to the target time. Positive values mean that the prediction is closer to the true earlier, healthier frame than simply copying the nearest observed diseased scan; negative values mean that the prediction is farther from the ground truth than this copy baseline.

The second diagnostic is the dynamics ratio,
\[
\mathrm{Dyn}
=
\frac{\mathbb{E}_t\|\hat{x}_{t+1}-\hat{x}_t\|}
{\mathbb{E}_t\|x_{t+1}-x_t\|},
\]
defined as predicted temporal change divided by ground-truth temporal change. A value near $0$ indicates temporal collapse to a static prediction, $1$ indicates correctly scaled temporal change, and values above $1$ indicate overshoot.

\section{Results}
\subsection{Synthetic Data}
\begin{figure}[H]
\centering
\includegraphics[width=0.99\textwidth]{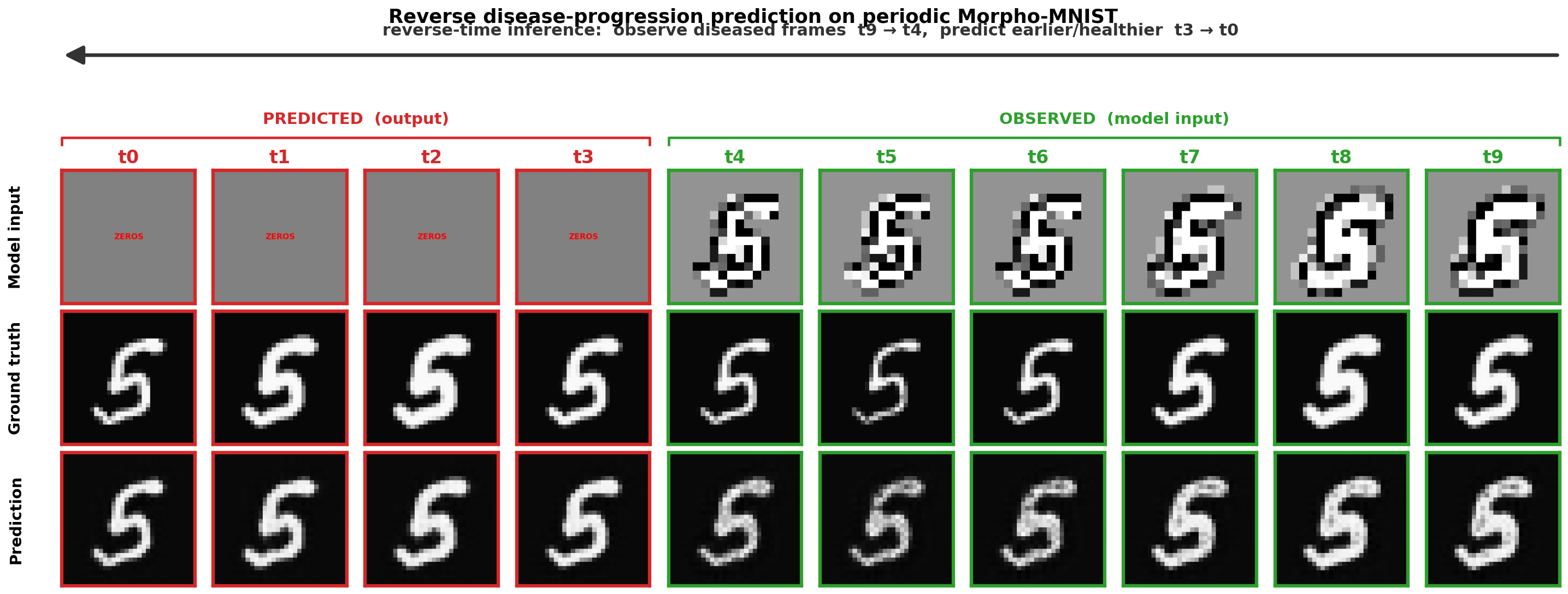}\\[3pt]
\caption{Visual results on Morpho-MNIST periodic reverse prediction. Given only diseased late frames ($t=4$--$9$, green), the model reconstructs healthier earlier frames ($t=0$--$3$, red). Rows show the masked input, ground truth, and prediction.}
\label{fig:morpho}
\end{figure}

Figure~\ref{fig:morpho} shows qualitative reverse prediction results on the simulated periodic Morpho-MNIST benchmark. Given only the diseased late frames ($t=4$--$9$), the model reconstructs the earlier healthier frames ($t=0$--$3$) and produces non-zero per-step changes, recovering both the thinning and thickening phases of the underlying non-monotonic trajectory. Quantitatively, predicted per-frame thickness tracks the analytic sine trajectory with Pearson $r=0.965$, close to the ground-truth-versus-sine correlation of $r=0.984$. An undertrained checkpoint gives only $r=0.039$, confirming that the signal comes from learned dynamics rather than the codebook alone. Over 100 samples and two seeds, reverse prediction reaches MSE $0.00561\pm0.00040$, PSNR $20.27\pm0.16$, and SSIM $0.5361\pm0.0013$, improving MSE over copy-nearest by $74.8\%$ and over mean-observed by $35.1\%$. The bottom row of Fig.~\ref{fig:morpho} also shows that the Stage-1 reconstruction ceiling is much lower than the temporal error on this clean data (direct encode--decode MSE $0.00152$), indicating that the limiting factor is the inverse temporal problem rather than representation fidelity.

\subsection{Registered ADNI Results}
\begin{figure}[H]
\centering
\includegraphics[width=\textwidth]{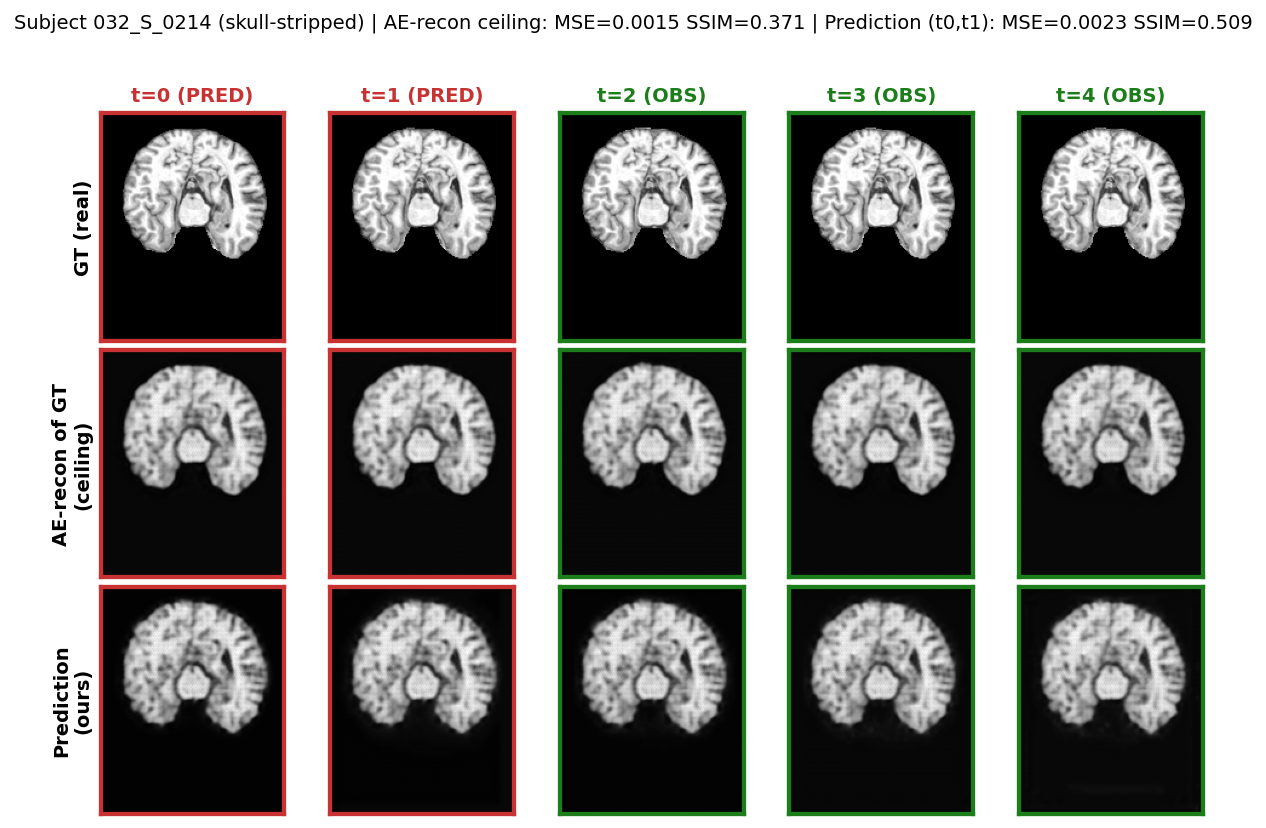}
\caption{Visual results on registered $1\,$mm ADNI. The frozen autoencoder already smooths the ground truth, and the Neural ODE prediction inherits this limitation. Green borders denote observed frames; red borders denote predicted frames.}
\label{fig:ae_ceiling_mni}
\end{figure}

Table~\ref{tab:adni} summarizes the primary $1\,$mm registered ADNI evaluation. Against raw pixel baselines, the model cannot win on voxel MSE: copy-nearest and mean-observed directly reuse observed voxels and avoid the autoencoder round trip, while every prediction from our model inherits the Stage-1 reconstruction ceiling. Figure~\ref{fig:ae_ceiling_mni} illustrates this limitation visually: the frozen autoencoder reconstruction is already smoothed before any temporal model is applied. Quantitatively, the Stage-1 ceiling has MSE $0.00131$, already above the pixel baselines' $0.00071$--$0.00088$. In the fair VQ-routed comparison, however, the Neural ODE gives the best perceptual SSIM ($0.574$ vs. $0.454$ and $0.470$), even though it still loses on MSE. The dynamics ratio is $1.29\pm0.23$: the model is not static, but it overshoots anatomical change. Per-stratum diagnostics show similar dynamics for stable CN, stable MCI, and MCI-to-dementia progressors, with uniformly negative disease-reversal scores; the learned dynamics are therefore real but not yet disease-specific.

\begin{table}[H]
\centering
\caption{Registered $1\,$mm ADNI diagnostics on predicted frames for 25 test subjects. Pixel baselines are shown for context but skip the autoencoder. Best fair SSIM is bold. Dyn. denotes dynamics ratio; Rev. denotes disease-reversal score.}
\label{tab:adni}
\footnotesize
\begin{minipage}{.58\linewidth}
\centering
\begin{tabular}{@{}lcc@{}}
\toprule
Method & MSE $\downarrow$ & SSIM $\uparrow$ \\
\midrule
Neural ODE & $0.00238$ & $\mathbf{0.574}$ \\
VQ copy-nearest & $0.00168$ & $0.454$ \\
VQ mean-observed & $0.00157$ & $0.470$ \\
\midrule
Pixel copy-nearest & $0.00088$ & $0.957$ \\
Pixel mean-observed & $0.00071$ & $0.963$ \\
\bottomrule
\end{tabular}
\end{minipage}\hfill
\begin{minipage}{.38\linewidth}
\centering
\begin{tabular}{@{}lcc@{}}
\toprule
Stratum & Dyn. & Rev. \\
\midrule
Overall & $1.29$ & $-1.88$ \\
Stable CN & $1.31$ & $-1.93$ \\
Stable MCI & $1.34$ & $-1.94$ \\
MCI prog. & $1.27$ & $-1.75$ \\
\bottomrule
\end{tabular}
\end{minipage}
\end{table}


On unregistered $64^3$ ADNI, the same model appears to beat baselines, improving MSE by $35.1\%$ over copy-nearest ($0.0218$ vs. $0.0336$) and $9.4\%$ over mean-observed ($0.0240$), with positive disease reversal in every stratum. This apparent success is largely a misalignment artifact: unregistered visits are not anatomically voxel-aligned, so pixel-copy baselines are penalized by pose and field-of-view differences. After $1\,$mm MNI registration, copy-nearest becomes close to optimal and the residual disease signal falls near or below the autoencoder noise floor.

\subsection{Ablations and Practical Findings}
There are two practical observations. First, single-scan inference remains viable on the unregistered $64^3$ pipeline: with only $t=4$ observed, the model predicts $t=0$--$3$ with MSE $0.0235$, beating a trivial single-scan copy baseline (MSE $0.0373$) by $37.0\%$. Adding more observed frames helps only modestly. Second, increasing Stage-1 capacity from a 32-code autoencoder without perceptual loss to a 256-code perceptual autoencoder worsens all reverse-prediction metrics. The richer autoencoder reintroduces high-frequency texture that the small temporal model cannot predict, so the compact discrete latent is preferable. For $1\,$mm whole-brain training, Stage-1 memory scales nearly linearly at about $7.9\,$GB per million voxels; the skull-stripped brain ($160\times188\times152$, $4.57$M voxels) fits at $36.2\,$GB on a single $48\,$GB GPU, whereas a skull-on head ($\sim7.5$M voxels, $\sim59\,$GB) does not.


\section{Discussion, limitations and Conclusion}
On synthetic non-monotonic data based on Morph-MNIST, the Neural ODE learns clear reverse dynamics. On registered ADNI, two constraints dominate. First, the frozen autoencoder sets a fidelity ceiling: direct reconstruction of a target frame already smooths fine anatomy, so no decoded temporal prediction can recover detail absent from the decoder. Second, after spatial standardization, three-year inter-visit change is close to the autoencoder noise floor. This explains why the model can be perceptually preferred in the fair comparison while losing voxel MSE to raw pixel copying.

These findings make reverse prediction a useful diagnostic benchmark, not yet a clinically validated disease-reversal system. Limitations include the modest ADNI test set, few progressors, deterministic trajectories, dynamics that are not yet disease-specific, and no ROI-level biomarker validation. Future work should raise the fidelity ceiling with refinement decoders or diffusion-based super-resolution, model uncertainty over plausible past trajectories, and condition dynamics on diagnosis or biomarkers while preserving strict spatial standardization. 

In summary, reverse disease progression prediction is a promising counterfactual synthesis setting for patient-specific progression analysis, but trustworthy claims require aligned data, representation-matched baselines, and explicit reporting of reconstruction ceilings. We hope this work provides insights into developing deep learning models to discover incubation periods for early detection and intervention.

\end{document}